\documentclass{article}
\usepackage{ijcai19}

\usepackage{times}
\usepackage{soul}
\usepackage{url}
\usepackage[hidelinks]{hyperref}
\usepackage[utf8]{inputenc}
\usepackage[small]{caption}
\usepackage{graphicx}
\usepackage{subfigure}

\usepackage{amsmath}
\usepackage{booktabs}
\usepackage{algorithm}
\usepackage{algorithmic}
\usepackage{amsfonts}
\title{3D Dense Separated Convolution Module for Volumetric Image Analysis}

\iftrue
\author{
Lei Qu$^1$
\and
Changfeng Wu$^2$\and
Liang Zou$^{3}$
\affiliations
$^1$Anhui University\\
$^2$Anhui University\\
$^3$China University of Mining and Technology\\
\emails
qulei@ahu.edu.cn,
ahwcf1995@163.com,
liangzou@ece.ubc.ca,
}
\fi

\begin{document}

\maketitle

\begin{abstract}
With the thriving of deep learning, 3D Convolutional Neural Networks have become a popular choice in volumetric image analysis due to their impressive 3D contexts mining ability. However, the 3D convolutional kernels will introduce a significant increase in the amount of trainable parameters. Considering the training data is often limited in biomedical tasks, a tradeoff has to be made between model size and its representational power. To address this concern, in this paper, we propose a novel 3D Dense Separated Convolution (3D-DSC) module to replace the original 3D convolutional kernels. The 3D-DSC module is constructed by a series of densely connected 1D filters. The decomposition of 3D kernel into 1D filters reduces the risk of over-fitting by removing the redundancy of 3D kernels in a topologically constrained manner, while providing the infrastructure for deepening the network. By further introducing nonlinear layers and dense connections between 1D filters, the network’s representational power can be significantly improved while maintaining a compact architecture. We demonstrate the superiority of 3D-DSC on volumetric image classification and segmentation, which are two challenging tasks often encountered in biomedical image computing. 
\end{abstract}

\section{Introduction}

During the last few years, Deep Learning (DL) and especially Convolutional Neural Networks (CNNs) have revolutionized computer vision and set new standards for various challenging tasks, such as image classification and semantic segmentation. Since these tasks are also shared in diagnostics, pathology, high-throughput screening, cellular and molecular image analyzing and more, the thriving of deep learning was also witnessed in the field of biomedical image analysis~\cite{2}.

However, compared to 2D images mostly used in computer vision, image data encountered in biomedical field are often volumetric. The substantial difficulties in annotating and interpreting of 3D volumetric data generally result in a much smaller training set than that of computer vision. In addition, in order to effectively explore the 3D contexts which is essential in the volumetric data analysis, much effort has to be paid in the designing of network. Current efforts often leading to either a significant increasing in the amount of learnable parameters or the complexity in the network designing and training. When dealing with large 3D image volumes, the computational cost as well as memory requirements will also being damaging even with the cutting-edge hardware. Therefore, how to effectively explore the 3D contexts and train an efficient volumetric network with limited training data is still an open problem in volumetric image analysis. 

\begin{figure}
\centering
\includegraphics[width=8.5cm,height=3.3cm]{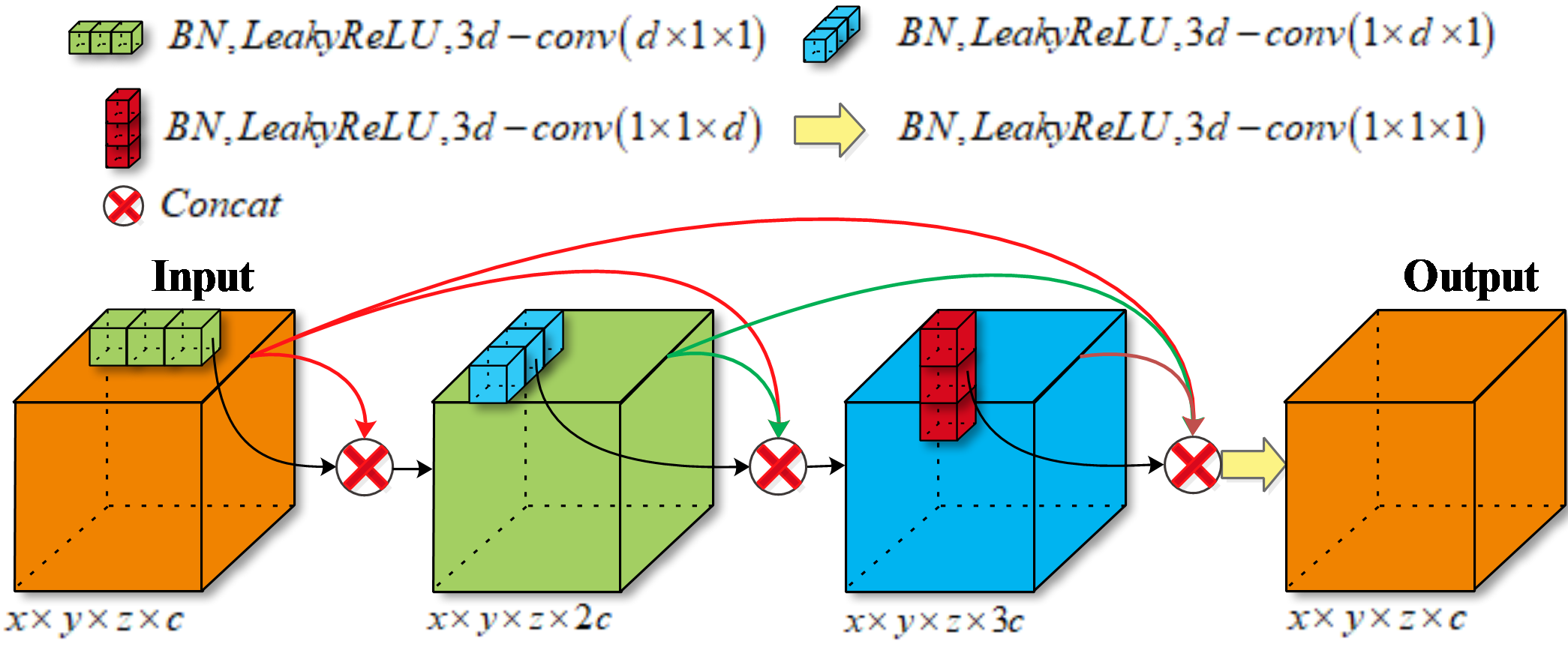}
\caption{Overview of 3D-DSC. For demonstration purpose, we just show one channel of the input and the feature volumes.}
\label{2}
\end{figure}

In order to process 3D volumes using CNNs, many schemes have been proposed in the past few years. One straightforward solution is to apply the conventional 2D CNNs on each volume slice separately~\cite{1}. Apparently, this method is a non-optimal use of the volumetric data since the contextual information along the third dimension is disregarded. To make a better use of the 3D context, the tri-planer schemes~\cite{Wolterink} suggested to applying 2D CNNs on three orthogonal planes (i.e., xy, xz and yz planes). Since the inter-slice information is utilized through a selective choosing of input data, only a small fraction of 3D information is explored~\cite{zheng2019a}. By viewing the adjacent volume slices as a time series, the recurrent neural network (RNN) was adopted to distil the 3D context from a sequence of abstracted 2D context~\cite{Chen2016Combining}. Due to the asymmetry nature of network design, the intra and inter-slice information cannot be treated and explored equally.

Currently, the 3D CNNs that take 3D convolution kernels as the basic unit~\cite{KS4ND}, and their hybrid with 2D CNNs~\cite{lee2015recursive}, have become the most popular choice in volumetric networks design. In addition to impressive 3D contexts mining ability, the popularity of 3D CNNs is also due to the simple structure nature of 3D operations (e.g. 3D convolutions, 3D pooling and 3D up-convolutions) and their similar usage to corresponding 2D operations. As a commonly adopted strategy, a 3D CNNs can be constructed from the modern 2D CNNs by replacing the 2D operations with their 3D counterparts.

However, the utilization of 3D operations, especially the 3D convolutional kernels, will introduce a huge increase in the amount of trainable parameters, as well as significant memory and computational requirements~\cite{Lai2015Deep}. Considering the limited training data often encountered in biomedical tasks, to avoid over-fitting, a tradeoff has to be made between the model size and its representational power. With these limitations, the existing 3D CNNs tending to contain much less layers than modern 2D CNNs. Since the impact of network’s depth has been extensively demonstrated to improve the performance in computer vision~\cite{11,12}, there was still much room to dig the potential of 3D CNNs and improve their representational power.

In this paper, instead of modifying the network’s overall architecture to circumvent the tradeoff between model size and its representational power, we address this dilemma by looking into the very basic unit of 3D CNNs---3D convolutional kernels, and proposing to replace them with a compact module that possesses better parameter efficiency and stronger nonlinear representational power. We named the proposed module 3D Dense Separated Convolution (3D-DSC), and Figure \ref{2}  illustrates its layout schematically.  

The 3D-DSC module is constructed by a series of densely connected 1D filters. The decomposition of 3D kernel into 1D filters allivating the risk of over-fitting by removing the redundancy within 3D kernels in a topologically constrained manner, while providing the infrastructure for deepening the network. The nonlinear layers inserted between 1D filters is responsible for the boosting of block’s nonlinearity as well as its representational power. The dense connections between 1D filters ensures efficient propagation of information and gradient flow, thus facilitate the training of deepened network. Finally, the 1$\times$1$\times$1 convolution attached in the end of block acts as a bottleneck layer to reduce the number of output feature volumes.  

Compared with direct 3D convolutions, 3D-DSC not only effectively deepens the network thus improving network’s representational power, but also considerably reduces the number of parameters. This feature is especially useful when training data is limited. Note that, our 3D-DSC is not limited to any specific architecture or application and it can be used to boost the performance by directly substituting the original 3D convolutional kernels.

\section{Related Work}

As the depth of network in computer vision has become saturated, research focusing on designing a more compact and parameter efficient architecture has received more attention recently. However, in biomedical image computing, there is still few effort be dedicated to this aspect. The present work mainly relies on the following efforts in the field of computer vision. 

On the way to improve parameter efficiency, CNNs~\cite{lecun} set the first milestone by introducing a parameter sharing mechanism. In the form of convolution, all neurons in a single depth slice of CNN are forced to share the same parameters, thus considerably reducing the overall number of  parameters. Another prominent design pattern is the bottleneck unit introduced in ResNet~\cite{11}. Constituted by two 1$\times$1 convolution layers, the bottleneck pattern explores the channel-wise redundancy in a shrink-and-expand manner. This idea was later adopted in Shufflenet~\cite{shufflenet} and Inception-v4~\cite{inception-v4} to reduce the computation and memory consumption. In~\cite{rethinking}, a spatial separation of the convolution operator was proposed, where the 3$\times$3 kernels were separated into two consecutive kernels of shapes 3$\times$1 and 1$\times$3. Recently MobileNet~\cite{mobilenets} took a step further and proposed depthwise separable convolutions. The resulting structure is many times more efficient in terms of memory and computation. The bypassing pattern, which was initially proposed in Highway Networks~\cite{highway} to facilitate the training of deeper networks. Its impressive parameter efficiency has also been discovered and confirmed with the extension of ResNet~\cite{11} and DenseNet~\cite{densely}.

Rather than designing a parameter efficient network, another group of works resort to exploring the redundancy of network in a post-processing manner. Among these efforts, the Low-Rank Approximation (LRA) methods are most relevant to ours. By viewing the convolutional layers as high-order tensors, these methods compress convolutional layers of pre-trained networks by finding their appropriate LRA. Using low-rank decomposition to accelerate convolution was first suggested by ~\cite{RigamontiLearning} in codebook learning. In the context of CNNs, ~\cite{dentonexploiting} proposed a canonical polyadic (CP) decomposition and clustering scheme for the convolutional kernels. Pre-trained 3D filters are approximated by a consecutive 1D filters and the error is minimized by using clustering and post training. ~\cite{jaderbergspeeding} suggested using different tensor decomposition schemes, an iterative schemes was employed to get an approximate local solution. ~\cite{lebedevspeeding} further extend the use of CP decomposition and propose a different low-rank architecture that enable both approximating an already trained network and training from scratch. Since LRA methods are mainly aim to speed up CNNs, and by approximating the weights of pretrained convolutional layers, the improvement of network’s nonlinear representational power are generally disregarded or even scarified.

\section{Methods}
We start this section by discussing the 3D separability of 3D convolutional kernels and the issues it may arise. Then, based on the infrastructure provided by the spatial decomposition of kernels, we detail the construction of our 3D-DSC module. 

\begin{figure}
\centering
\includegraphics[width=8.5cm,height=3.5cm]{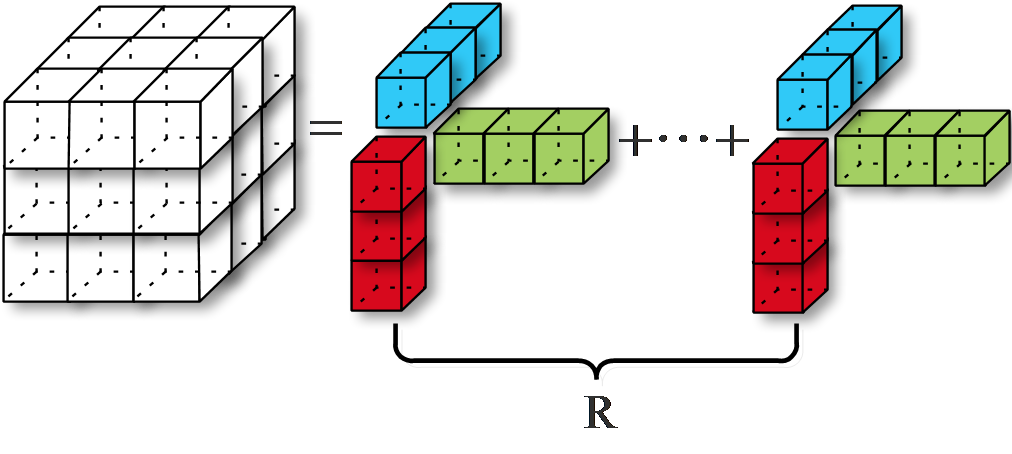}
\caption{Visualization of the separated convolution. The white cube is the original rank-R 3D convolutional kernel and the 1D kernels are the CP decomposition of the white cube.}
\label{1}
\end{figure}

\subsection{3D Separability of Convolutional Kernels}
Given a volumetric image, the 3D convolution operation with stride one can be formulated as bellow in an element-wise fashion: %
\begin{align}
    V_{jk}^l(x,y,z) =& \sum\limits_{x' = 1}^X {\sum\limits_{y' = 1}^Y {\sum\limits_{z' = 1}^Z {F_k^{l{\rm{ - }}1}(x - x',y - y',z - z')} } } \times \nonumber\\
    \times & W_{jk}^l(x',y',z')
\end{align}%
where $ W_{jk}^l $ is the 3D kernel of size $X \times Y \times Z$ in the {\it l}th layer which connected to the {\it k}th input feature volume $F_k^{l{\rm{ - }}1}$ in the previous layer and the {\it j}th output feature volume $F_j^l$, $W_{jk}^l(x',y',z')$ is the element-wise value of the 3D convolution kernel. Assume the {\it l}th layer has $K$ input feature volumes, and let $\sigma ( \cdot )$ denotes the element-wise non-linear activation function and $b_j^l$ the corresponding bias term, the output feature volume $F_j^l$ is obtained as:%
\begin{align}
    F_j^l = \sigma (\sum\limits_{k = 1}^K {V_{jk}^l + b_j^l} )
\end{align}%

Mathematically, the 3D kernel tensor $W_{jk}^l$ can be factorized into a linear combination of rank-one tensors according to the CP decomposition:
\begin{align}
W_{jk}^l = \sum\limits_{r = 1}^R {a_{jkr}^l \otimes b_{jkr}^l \otimes c_{jkr}^l}
\end{align}%
where $R$ is the rank of $W_{jk}^l$, $\otimes$ denotes the outer product operation and $a_{jkr}^l \in \mathbb{R}^X$, $b_{jkr}^l \in \mathbb{R}^Y$,  $c_{jkr}^l \in \mathbb{R}^z$ are 1D vectors. Element wisely, the above equation can be rewritten as:
\begin{align}
	W_{jk}^l(x,y,z) = \sum\limits_{r = 1}^R {a_{jkr}^l(x)b_{jkr}^l(y)c_{jkr}^l(z)}
\end{align}%

Substituting (4) into (1) gives the following equivalent expression for the evaluation of the 3D convolution:
\begin{align}
	V_{jk}^l(x,y,z) = &\sum\limits_{r = 1}^R {\left( {} \right.} \sum\limits_{z' = 1}^Z {} \left( {} \right.\sum\limits_{y' = 1}^Y {} \left( {} \right.\sum\limits_{x' = 1}^X {} F_k^{l{\rm{ - }}1}(x - x',y - y', \nonumber\\
	&,z - z')a_{jkr}^l(x')\left. \right)b_{jkr}^l(y')\left. \right)c_{jkr}^l(z')\left. \right)
\end{align}%

With this formulation, the 3D convolution can be recasted as a sequence of 1D convolutions. From inside out, the calculation within the parenthesizes can be viewed as: first convolve the feature volume with a 1D filter $a_{jkr}^l$ along the X dimension, then followed with the 1D convolution with $b_{jkr}^l$ and $c_{jkr}^l$ along Y and Z dimension successively.

Assume the rank of kernel tensor $W_{jk}^l$ equal to one (i.e. R=1), the 3D convolution can be decomposed into a sequence of three 1D convolutions as shown in Figure \ref{1} ($R=1$). Note that convolution is a linear operator, the 1D filters as shown in Figure \ref{1} can be arranged in any order.

Rank-1 is a strong assumption and the intrinsic rank of $W_{jk}^l$ is generally higher than one in practice. The Equ. (3) shows that the rank-R tensor is the sum of R rank-1 tensors, this suggested that the rank-R topology can be constructed by simply concatenating R copies of rank-1 case as shown in Figure \ref{1}.

\subsection{ 3D Dense Separated Convolution Module}

Although the 3D  separated convolution topology described in the previous section is mathematically equivalent to direct 3D convolution, the profits of this decomposition are: 

First, the rank-constraints of 3D convolution kernels can be easily encoded in the network’s topology by stacking k (k$<$R) groups of 1D convolutions (as seen in Figure \ref{1}). Once the model structure is defined, we can leverage the traditional CNNs training method to learn more compact weights from scratch, thus avoiding the post-processing stage of LRA. In addition, the information loss and performance degradation caused by low-rank constraints can be minimized as a whole upon training. We will show that the precision can even be increased in our experiment section. 

Second, when a rank-k topology is applied to replace the original full rank 3D convolution kernel, the number of independent parameters per-filter can be reduced from $X \times Y \times Z$ to $(X+Y+Z) \times k$, which results in a significant reduction of overall learnable parameters for small k considering the huge number of filters deployed in the network. Since the training data size in many biomedical tasks is much smaller than that of computer vision, this reduction will alleviates the risk of over-fitting and further enable deeper network designing. 

Finally, the cascaded 1D convolutions naturally provides the infrastructure to further improve its nonlinear representation power. Since the linear combination of convolutions is still linear, the current decomposed topology can only increase the network’s visual depth but its effective depth. However, with this decomposed structure, the effective depth of network can be easily increased by inserting the nonlinear activation layers (e.g. LeakyReLU layers) between the concatenated 1D convolutions, thus increasing the nonlinearity of network and encouraging the learning of more discriminative features. 

However, there are two issues inherited in this kernel decomposition. First, the serialized model with 1D convolutions is more vulnerable to vanishing gradient problem than standard 3D CNNs. Accompanied with the increasing of the network’s depth, longer gradient propagation paths may result in fast gradient decaying as well as difficulty of optimization. Second, once the nonlinear activation layer is inserted between the 1D filters, the different ordering of the 1D filters will no longer be equivalent.
Inspired by the recent success of densely connected networks~\cite{densely}, we propose to extend the 3D separated convolution discussed in the previous section by further introducing dense connections between 1D filters. Figure \ref{2} illustrates the layout of resulting rank-R 3D Dense Separated Convolution (3D-DSC) module schematically. 

Similar to the densenet, we introduce direct connections from any layer to all subsequent layers within each block. In order to maximize the information flow, the features are concatenated, and then followed with a composited operations including Batch Normalization (BN) and leaky rectified linear units (LeakyReLU) before they are passed to the next layer. In our implementation, we restrict each layer to produce the same number of feature volumes as input. Assume there are k feature volumes in the input layer, the concatenate operation after the last 1D convolution layer will accumulate the feature volume to the number of 4k. In order to make the number of output feature volume consistent with that of direct 3D convolution, an additional bottleneck layer is appended behind the last 1D convolution layer. With this designing, the extension from rank-1 3D-DSC to the rank-k case will be same as the method discussed in previous section, i.e., by simply stacking k copies of the rank-1 topology.  

By introducing the within block dense connections, each 1D kernel are provided with the opportunity to directly access the input feature volume, thus to some extent alleviate their ordering problem. In addition, the employment of dense connections also bring the three following benefits that relief our previous concerns in a point-by-point manner. First, direct connections between all layers help improving the flow of information and gradients through the network, alleviating the problem of vanishing gradient. Second, short paths to all the feature volumes in the architecture introduce an implicit deep supervision. Third, dense connections have a regularizing effect, and considering the reduction in the number of parameters introduced by 3D-DSC, such a joint effort will substantially reduce the risk of over-fitting under limited training data.  

Since the size and number of feature volume of our 3D-DSC module is consistent with that of direct 3D convolution, we can directly substitute the 3D convolution with 3D-DSC and enjoy the its benefits. If using a high-level library such as Keras or TensorFlow-Slim, it will takes only several lines of code.

\begin{table}
\centering
\begin{tabular}{lll}
\toprule
$A_n$       & $B_n$  & $C_n$     \\
\hline
\multicolumn{3}{c}{Input(3D multi-channel MRI)} \\
\hline
3D-conv3-32       & 3D-conv3-32  & 3D-conv3-32 \\
3D-conv3-32       & 3D-conv3-32  & 3D-conv3-32 \\
\hline
\multicolumn{3}{c}{Maxpooling} \\
\hline
3D-conv3-64       & 3D-conv3-64  & 3D-conv3-64 \\
3D-conv3-64       & 3D-conv3-64  & 3D-DSC-64 \\
\hline
\multicolumn{3}{c}{Maxpooling} \\
\hline
3D-conv3-128       & 3D-conv3-128  & 3D-conv3-128 \\
3D-conv3-128$\times n$    &3D-DSC-128$\times n$  &3D-DSC-128$\times n$ \\
\hline
\multicolumn{3}{c}{Maxpooling} \\
\hline
3D-conv3-256       & 3D-conv3-256  & 3D-conv3-256 \\
3D-conv3-256$\times n$    &3D-DSC-256$\times n$  &3D-DSC-256$\times n$ \\
\hline
\multicolumn{3}{c}{Maxpooling} \\
\hline
3D-conv3-512       & 3D-conv3-512  & 3D-conv3-512 \\
3D-conv3-512$\times n$    &3D-DSC-512$\times n$  &3D-DSC-512$\times n$ \\
\hline
\multicolumn{3}{c}{Global-Average-Pooling} \\
\hline
\multicolumn{3}{c}{3D-conv1-2} \\
\hline
\multicolumn{3}{c}{Soft-max} \\
\bottomrule
\end{tabular}
\caption{Architecture overview (shown in columns). $A_n$ are the networks with normal 3D convolution, $B_n$ and $C_n$ are the networks with 3D-DSC, and $n$ represents the number of the additional convolution layers. The 3D convolutional kernel parameters are expressed as “3D-conv$\left\langle {\rm{kernel\;size}} \right\rangle $/3D-DSC-$\left\langle {\rm{number\;of\;output\;channel}} \right\rangle $”. The LeakyReLU activation layer and Batch Normalization layer are not shown here for brevity.}
\label{tab:1}
\end{table}

\section{Experiments and Results}
In this section, we evaluate the proposed module on two different volumetric image analysis tasks (Attention Deficit Hyperactivity Disorder Diagnosis and Brain Tumor Segmentation) with comparison to several state-of-the-arts methods. In addition to the precision evaluation, the components, depth and overfitting analyses are also provided to illustrate the effectiveness and superiority of our method.

\subsection{Attention Deficit Hyperactivity Disorder Diagnosis}
Attention Deficit Hyperactivity Disorder Diagnosis (ADHD) is one of the most common mental-health disorders, affecting around 5\%-10\% of school-age children. In order to automatically diagnose this disorder, MR images, including structure MRI (sMRI) and functional MRI (fMRI), have been investigated in many studies. The MRI data analyzed in this paper is from the ADHD200 consortium~\cite{45}. Initially, they post a large training dataset including 776 samples comprised of 491 typically developing individuals and 285 patients with ADHD. For each sample, both fMRI scans and associated T1weighted structural scans are provided. Besides, three kinds of voxel-based morphometric features, including gray matter (GM), white matter (WM) and cerebrospinal fluid (CSF) are also provided in~\cite{46}. In our experiment, these features are regarded as three individual input channels of network.

\begin{figure}
\centering
\includegraphics[height=4.84cm,width=6.4cm]{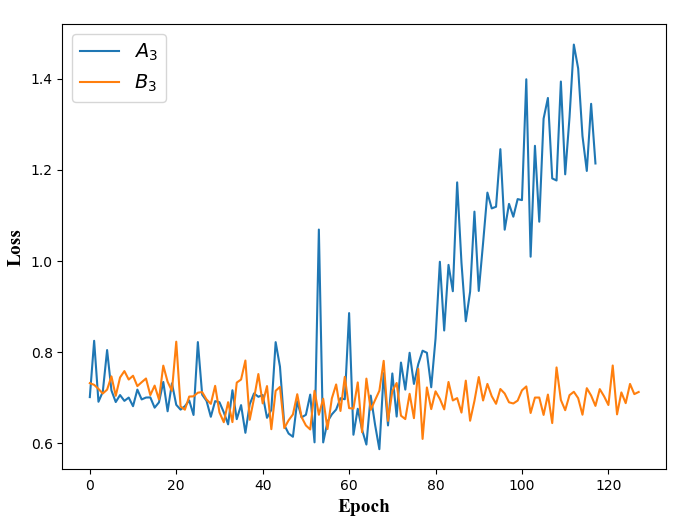}
\caption{The validation loss of $A_3$ and $B_3$ network. We do not employ dense connection or extra activation layer during the separated convolution in the $B_3$ network.}
\label{3}
\end{figure}

\paragraph{Network Architecture.} Table \ref{tab:1} shows the configurations of the baseline models ($A_n$) and their 3D-DSC enhanced versions ($B_n$ and $C_n$). All networks start with two 3D convolutional layers and one pooling layer.The difference between $B_n$ and $C_n$ is whether 3D-DSC modules are used or not between first two pooling layers. Starting from the second maxpooling layer, both $B_n$ and $C_n$ are constructed by repeating a combination of one 3D convolution layer, n 3D-DSC layers and one pooling layer. Then, the global average pooling layer and $1\times1\times1$ convolution are applied on the feature volumes, and SoftMax is employed as the last layer for classification.

\paragraph{Training and evaluation setting.} We employ the cross-entropy as the loss function. It is worth noting that adaptive optimization methods have better performance in the early stage of training but are outperformed by SGD at later stages. To minimize the effect of random initialization, we firstly train the model with random initialization and Adam optimizer; then, we refine the model with SGD optimizer. The Early-Stopping strategy is used with patience of 50. We denote the mean difference between the training loss and the validation loss within the last 50 epochs as Overfitting Distance (OD), which can be used to evaluate the ability of network in coping with overfitting. In our experiments, we employ 5-fold cross validation to evaluate the proposed method.

\begin{table}
\centering
\begin{tabular}{lrr}  
\toprule
Network  &Accuracy &OD \\
\midrule
$A_0$       & 73.17\%  & 0.2785      \\
$A_1$       & 74.89\%  & 0.2806      \\
$A_2$       & 74.53\%  & 0.2807      \\
$A_3$       & 71.94\%  & 0.2961      \\
$B_3$       & 75.22\%  & {\bf 0.2526}      \\
$B_6$       & 75.74\%  & 0.2809      \\
$C_6$       & {\bf 76.70}\%  & 0.2580      \\
\bottomrule
\end{tabular}
\caption{Performance comparison based on 5-fold cross validation. ${A_{\rm{0}}} \sim {A_3}$  are normal 3D CNNs with different depth. ${B_3}$, ${B_6}$ and ${C_6}$ are the separated 3D CNNs with 3D-DSC. {\bf OD} denotes the overfitting distance.}
\label{tab:3}
\end{table}

\paragraph{Accuracy and analysis of network's depth.}It is well known that the depth of network has big impact on its performance. Table \ref{tab:3} shows the accuracy and OD score of network with different depth configurations. For the baseline method ($A_n$), we can see that $A_1$ achieves the best result. However, its performance will decline as the network deepens. We believe that the aggravation of overfitting is responsible for this degradation since the number of parameter will increased dramatically with deeper network. In contrast, we can observe a stable performance improvement with our 3D-DSC enhanced versions even when its depth reaches $B_6$ and $C_6$. Interestingly, note that $B_6$ and $A_2$ have the similar number of parameters. These results confirm deeper network and be obtained and effectively trained with 3D-DSC, thus improving the network’s representational power. Moreover, as shown in Table \ref{tab:4}, compared with several state-of-the-arts methods attempting to assist the diagnosis, $C_6$ outperforms the others with large margin on the ADHD-200 even if only single modality of dataset is used in our method.

\begin{table}
\centering
\begin{tabular}{lllrr}  
\toprule
Network  &DC &Activation &Accuracy &OD \\
\midrule
$B_3$       & no  &no   &74.26\%  &0.2733   \\
$B_3$       & no  &yes    &74.83\% &0.2638  \\
$B_3$       &yes  &yes    &{\bf75.22}\% &{\bf0.2580}  \\
\bottomrule
\end{tabular}
\caption{Ablation studies for applying dense connection (DC) and activation layer (Activation for abbreviation) in the proposed 3D-DSC. {\bf OD}: overfitting distance. }
\label{tab:2}
\end{table}

\begin{table}
\centering
\begin{tabular}{llr}  
\toprule
Method  &Classifier &Accuracy \\
\midrule
~\cite{47}       & MKL  & 61.54\%      \\
%~\cite{48}       & SVM  &62.57\%      \\
%~\cite{49}       & SVM  & 62.81\%      \\
~\cite{410}     & SVM  &63.57\%      \\
~\cite{411}     & SM 3D CNN  &66.04\%      \\
~\cite{411}     & MM 3D CNN  & 69.15\%      \\
3D-DSC$(C_6)$       & SM 3D CNN  &{\bf 73.68}\%      \\
\bottomrule
\end{tabular}
\caption{Diagnosis performance comparisons between the proposed method and state-of-the-art methods based on the ADHD-200 dataset. {\bf MKL}: multi kernel learning. {\bf SVM}: support vector machine. {\bf SM}: single modality. {\bf MM}: multi modalities.}
\label{tab:4}
\end{table}

\paragraph{Ablation studies.} To investigate the effect of nonlinear activation layers and dense connections inserted between the separated 1D filters, we report the performances of $B_3$ with and without nonlinear layers and dense connections in Table \ref{tab:2}. We can see that both of them contribute to the performance improvement, and the best result can be obtained by a combination of them.

\paragraph{Overfitting.} To further confirm the ability of 3D-DSC in coping with overfitting, we compare the performance of baseline method with ours by removing the Batch Normalization (BN) layers. For demonstration purposes, we set the batch-size to 1, the learning rate is initially set to 0.0001 and decreases by a factor of 10 when the validation error stops decreasing. Figure ~\ref{3} shown the loss curve of $A_3$ and $B_3$ (without BN layer) on the validation dataset. We can see that the validation loss of $A_3$ increase rapidly since 60 epochs, while that of $B_3$ network remains stable, even after 100 epochs. Furthermore, the OD score of $A_3$ is 0.9962, which is significantly larger than 0.2946 of $B_3$. Compared with $A_3$, the more compact structure and much less learnable parameters of $B_3$ make it less susceptible to overfitting.

\subsection{The Brain Tumor Segmentation on BRATS 2017}
In this section, we evaluate the proposed method on another challenging task of brain tumor segmentation, using the public available dataset of BRATS 2017 challenge~\cite{412}. The training dataset contains 285 multisequence MRI of patients diagnosed with low-grade gliomas or high-grade gliomas. In each case, four MR sequences are available: T1, T1+gadolinium, T2 and FLAIR. In our experiment, we resize all volumes to $(64 \times 64 \times 64)$ and four MRI sequences for each sample are combined as a multichannel volume as input. Three state-of-the-arts methods including 3D U-net~\cite{cicek20163d} (along with its dropout and stride 2 convolution enhanced versions), V-net~\cite{414} and method proposed in~\cite{413} are evaluated and compared.

\begin{figure}
\centering
\subfigure[]{\label{41}
\includegraphics[width=1.5cm]{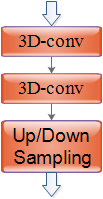}
}
\quad
\subfigure[]{\label{42}
\includegraphics[width=1.5cm]{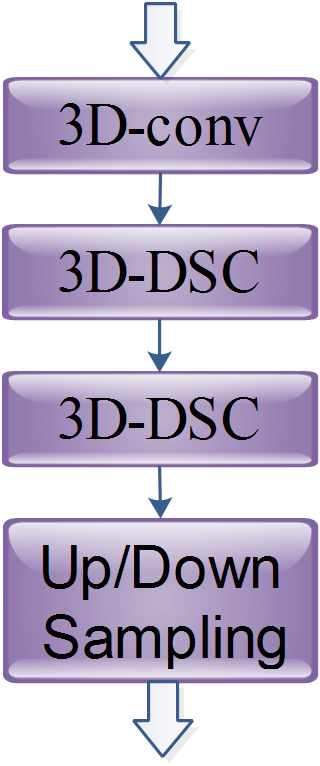}
}
\caption{The basic block of 3D U-net (shown in subfigure (a)) and the basic block of the network with 3D-DSC (shown in subfigure (b)), where we replace the second normal 3D convolution with 2 3D-DSC.}
\label{4}
\end{figure}

\begin{figure}
\centering
\subfigure[Ground truth]{
\includegraphics[width=1.6cm]{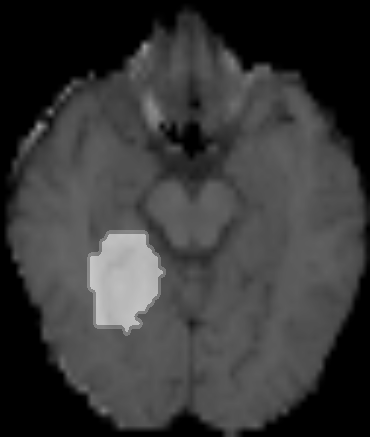}
}
\quad
\subfigure[3D U-net]{
\includegraphics[width=1.6cm]{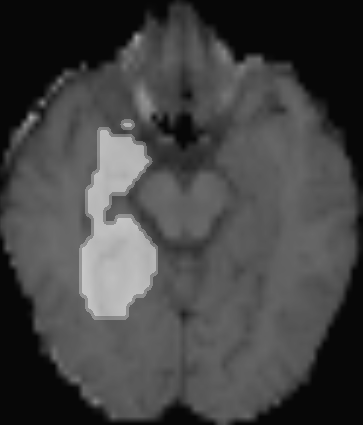}
}
\quad
\subfigure[V-net]{
\includegraphics[width=1.6cm]{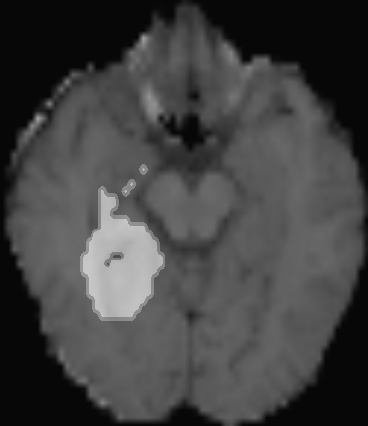}
}
\quad
\subfigure[ours]{
\includegraphics[width=1.6cm]{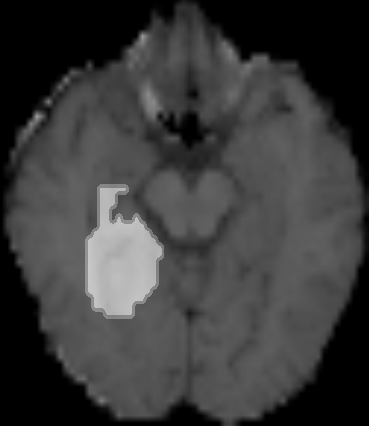}
}
\caption{ Example segmentation results on the BRATS 2017 Challenge dataset. From left to right: the ground truth, the segmentation result of 3D U-net, the segmentation result of V-net and the segmentation result of the proposed method.}
\label{5}
\end{figure}

5-fold cross-validation strategy is employed in this experiment. In each fold, 228 samples are used for training and 57 samples are used for validation. Our 3D-DSC enhanced version is based on 3D U-net. We replace the original 3D U-net block as shown in Figure \ref{4}.a by our 3D-DSC enhanced version as shown in Figure \ref{4}.b. Note that only the second 3D convolution in 3D U-net is substituted by two consecutive 3D-DSC modules, and the overall architecture of network remain unchanged. The Dice score obtained by different methods are shown in Table \ref{tab:5}. We can see that our method achieved the best performance with Dice score 0.7932, which outperforms the others with a large margin. It is worth noting that we did not adopt any trick in our method, such as dropout, replacing pooling with stride 2 convolution (s2-conv) and so on, although these tricks can slightly improve the performance of 3D U-net as reported in Table \ref{tab:5}. The qualitative segmentation results of different methods are presented in Figure \ref{5}, we can see that fine details can be better recovered by our method.

\begin{table}
\centering
\begin{tabular}{lr}  
\toprule
Method  &Dice score\\
\midrule
3D U-net       & 0.7554\%        \\
3D U-net(Dropout)       & 0.7592\%       \\
3D U-net(s2-conv)       & 0.7593\%       \\
~\cite{413}       & 0.7655\%       \\
V-net~\cite{414}       & 0.7685\%       \\
3D U-net(3D-DSC)       & {\bf 0.7932}\%       \\
\bottomrule
\end{tabular}
\caption{The experimental results of the proposed method and state-of-the-art methods. We train and evaluate these methods with the same strategy on the BRATS2017 dataset.}
\label{tab:5}
\end{table}

\section{Conclusion and Discussion}
The effective and efficient exploration of 3D contextual information is essential in volumetric data analysis. Although the performance of CNNs in 2D image analysis is impressive, the predictive power of its 3D generalization (i.e., 3D CNNs) is always constrained by the number of samples, especially in biomedical image analysis. Considering the conflict between huge amount of parameters to learn in 3D CNNs and limited training samples which would quickly lead to overfitting, in this paper, we propose a novel 3D-DSC module to replace the traditional 3D convolutional kernels. The proposed 3D-DSC module consists of a series of densely connected 1D filters. This architecture is able to remove the redundancy within 3D kernels, while providing spaces for deepening the network, and therefore can effectively reduce the risk of overfitting. In addition, inspired by the recent success of residual network and densely connected networks, we extend the 3D separated convolution block by introducing dense connections within and between blocks. The dense connection provides an effective way to combine subsequent-layers and facilitates the flow of information. Furthermore, we investigate the effect of nonlinear activation layers between the concatenated 1D filters, which have the potentiality to increase representational power of the network and facilitate the learning of discriminative features. Experimental results on the ADHD classification and brain tumor segmentation demonstrate the superiority of the proposed 3D-DSC on volumetric image analysis. Note that 3D-DSC is not limited to any specific architecture or application and can be used to boost the performance by directly substituting the original 3D convolutional kernels. 

\bibliographystyle{named}
\bibliography{ijcai19}

\end{document}